\documentclass{llncs}
\usepackage{graphicx}
\usepackage{float}
\usepackage{subfigure}
\usepackage{booktabs}
\usepackage{hyperref}
\usepackage{rotating} 
\usepackage{caption}
\usepackage{multirow}
\begin{document}

\title{The CL-SciSumm Shared Task 2018:\\Results and Key Insights}
\titlerunning{CL-SciSumm 2018}  
%
\author{Kokil Jaidka\inst{1} \and Michihiro Yasunaga\inst{2} \and Muthu Kumar Chandrasekaran\inst{3} \\ Dragomir Radev\inst{2} \and  Min-Yen Kan\inst{3,4}}
\authorrunning{Kokil Jaidka et al.} 
\institute{University of Pennsylvania, USA
\and
Department of Computer Science, Yale University, USA
\and
School of Computing, National University of Singapore, Singapore
\and
Smart Systems Institute, National University of Singapore, Singapore \\
\email{\tt{jaidka@sas.upenn.edu}}
}

\maketitle              


\begin{abstract}
This overview describes the official results of the CL-SciSumm Shared
Task 2018 -- the first medium-scale shared task on
scientific document summarization in the computational
linguistics~(CL) domain. This year, the dataset comprised 60 annotated sets of
citing and reference papers from the open access research papers in
the CL domain. The Shared Task was organized as a
part of the $41^{st}$ Annual Conference of the Special Interest Group
in Information Retrieval (SIGIR), held in Ann Arbor, USA in July 2018.
We compare the participating systems in terms of two evaluation
metrics. The annotated dataset and evaluation scripts can be accessed and used by the community from:
\url{https://github.com/WING-NUS/scisumm-corpus}.
\end{abstract}


\section{Introduction}
\label{s:intro}
CL-SciSumm explores summarization of scientific research in the domain of computational
linguistics. The Shared Task dataset comprises the set of citation sentences (i.e., ``citances'') that
reference a specific paper as a (community-created) summary of a topic or
paper~\cite{qazvinian2008scientific}. Citances for a reference paper are considered a
synopses of its key points and also its key contributions and importance within an
academic community~\cite{nakov2004citances}. The advantage of using citances is that they
are embedded with meta-commentary and offer a contextual, interpretative layer to the
cited text. Citances offer a view of the cited paper which could complement the reader's
context, possibly as a scholar~\cite{sparckjones2007automatic} or a writer of a
literature review~\cite{jaidka2013deconstructing}.

The CL-SciSumm Shared Task is aimed at bringing together the summarization community to
address challenges in scientific communication summarization. It encourages the incorporation of new kinds of information in automatic scientific paper summarization, such as the facets of research information being summarized in the research paper, and the use of new resources, such as the mini-summaries written in other papers by other scholars, and concept taxonomies developed for computational linguistics. Over time, we anticipate
that the Shared Task will spur the creation of other new resources, tools, methods and evaluation
frameworks.

CL-SciSumm task was first conducted at TAC 2014 as part of the larger BioMedSumm Task\footnote{http://www.nist.gov/tac/2014}. It was organized in 2016 \cite{jaidka2017insights} and 2017 \cite{jaidka2017overview} as a part of the Joint Workshop on Bibliometric-enhanced Information Retrieval and Natural Language Processing for Digital Libraries (BIRNDL) workshop~\cite{mayr16:_editorial} at the  Joint Conference on Digital Libraries (JCDL\footnote{http://www.jcdl2016.org/}) in 2016, and the annual ACM Conference on Research and Development in Information Retrieval (SIGIR\footnote{\url{http://sigir.org/sigir2017/}}) in 2017 \cite{MayrCJ17}. This paper provides the results for the CL-SciSumm 2018 Task being held as part of the BIRNDL 2018 workshop at SIGIR 2018 in Ann Arbor, Michigan.

\section{Task}

CL-SciSumm defines two serially dependent tasks that participants could attempt, given a canonical training and testing set of papers. \\

\label{s:task}
\noindent \textbf{Given}: A topic consisting of a Reference Paper (RP) and ten or more Citing Papers (CPs) that all contain citations to the RP. In each CP, the text spans (i.e., citances) have been identified that pertain to a particular citation to the RP.
Additionally, the dataset provides three types of summaries for each RP:
\begin{itemize}
	\item {the abstract, written by the authors of the research paper.}
    \item {the community summary, collated from the reference spans of its citances.}
  	\item {a human-written summary, written by the annotators of the CL-SciSumm annotation effort.}
\end{itemize}

\noindent \textbf{Task 1A}: For each citance, identify the spans of text (cited text
spans) in the RP that most accurately reflect the citance. These are of the
granularity of a sentence fragment, a full sentence, or several consecutive
sentences (no more than 5). \\

\noindent \textbf{Task 1B}: For each cited text span, identify what facet of the paper
it belongs to, from a predefined set of facets. \\

\noindent \textbf{Task 2}: Finally, generate a structured summary of the RP from the
cited text spans of the RP. The length of the summary should not exceed 250
words. This was an optional bonus task.

\section{Development}
\label{s:dev}
The CL-SciSumm 2018 corpus comprises a training set that is randomly sampled research papers (Reference papers, RPs) from the ACL Anthology corpus and the citing papers (CPs) for those RPs which had at least ten citations. The prepared dataset then comprised annotated citing sentences for a research paper, mapped to the sentences in the RP which they referenced. Summaries of the RP were also included. The CL-SciSumm 2018 corpus included a refined version of the CL-SciSumm 2017 corpus of
40 RPs as a training set, in order to encourage teams from the previous edition to participate. For details of the general procedure followed to construct and annotate the CL-SciSumm corpus, the changes made to the procedure in CL-SciSumm-2016 and the refinement of the training set in 2017, please see
\cite{jaidka2017overview}.

The test set was an additional corpus of 20 RPs which was picked out of the ACL Anthology Network corpus (AAN), which automatically identifies and connects the citing papers and citances for each of thousands of highly-cited RPs. Therefore, we expect that that characteristics of the test set could be somewhat different from the training set. For this year's test corpus, every RP and its citing papers were annotated three times by three independent annotators, and three sets of human summaries were also created.

\subsection{Annotation}
\label{ss:annot}

The annotation scheme was unchanged from what was followed in previous
editions of the task and the original BiomedSumm task developed by
Cohen et. al\footnote{\url{http://www.nist.gov/tac/2014}}:
Given each RP and its associated CPs, the annotation group was instructed to find
citations to the RP in each CP. Specifically, the citation text, citation marker,
reference text, and discourse facet were identified for each citation of the RP
found in the CP.

\section{Overview of Approaches}
\label{s:methods}

Ten systems participated in Task~1 and a subset of three also participated in Task~2.
The following paragraphs discuss the approaches followed by the participating systems, in lexicographic order by team name.  \\

\textbf{System 2:} The team from the Beijing University of Posts and Telecommunications' Center for Intelligence Science and Technology \cite{system_2} developed models based on their 2017 system. For Task 1A, they adopted Word Mover’s Distance (WMD) and improve LDA model to calculate sentence similarity for citation linkage. For Task 1B they presented both rule-based systems, and supervised machine learning algorithms such as: Decision Trees and K-nearest Neighbor. For Task 2, in order to improve  the performance of summarization, they also added WMD sentence similarity to construct new kernel matrix used in Determinantal Point Processes (DPPs).

\textbf{System 4:} The team from Thomson Reuters, Center for Cognitive Computing \cite{system_4} participated in Task 1A and B. For Task 1A, they treated the citation linkage prediction as a binary classification problem and utilized various similarity-based features, positional features and frequency-based features. For Task 1B, they treated the discourse facet prediction as a multi-label classification task using the same set of features.

\textbf{System 6:}  The National University of  Defense Technology team \cite{system_6} participated in Task 1A and B. For Task 1A, they used a random forest model using multiple features. Additionally, they integrated random forest model with BM25 and VSM model and applied a voting strategy to select the most related text spans. Lastly, they explored the language model with word embeddings and integrated it into the voting system to improve the performance. For task 1B, they used a multi-features random forest classifier.

\textbf{System 7:} The Nanjing University of Science and Technology team (NJUST) \cite{system_7} participated in all of the tasks (Tasks 1A, 1B and 2). For Task 1A, they used a weighted voting-based ensemble of classifiers (linear support vector machine (SVM), SVM using a radial basis function kernel, Decision Tree and Logistic Regression) to identify the reference span. For Task 1B, they used a dictionary for each discourse facet, a supervised topic model, and XGBOOST. For Task 2, they grouped sentences into three clusters (motivation, approach and conclusion) and then extracted sentences from each cluster to combine into a summary.

\textbf{System 8:} The International Institute of Information Technology team \cite{system_8} participated in Task 1A and B. They treated Task 1A as a text-matching problem, where they constructed a matching matrix whose entries represent the similarities between words, and used convolutional neural networks (CNN) on top to capture rich matching patterns. For Task 1B, they used SVM with tf-idf and naive bayes features.

\textbf{System 9:} The Klick Labs team \cite{system_9} participated in Task 1A and B. For Task 1A, they explored word embedding-based similarity measures to identify reference spans. They also studied several variations such as reference span cutoff optimization, normalized embeddings, and average embeddings. They treated Task 2B as a multi-class classification problem, where they constructed the  feature vector for each sentence as the average of word embeddings of the terms in the  sentence.

\textbf{System 10:} The University of Houston team \cite{system_10} adopted sentence similarity methods using Siamese Deep learning Networks and Positional Language Model approach for Task~1A. They tackled Task~1B using a rule-based method augmented by WordNet expansion, similarly to last year.

\textbf{System 11:}  The LaSTUS/TALN+INCO team \cite{system_11} participated in all of the tasks (Tasks 1A, 1B and 2). For Task 1A, B, they proposed models that use Jaccard similarity, BabelNet synset embeddings cosine similarity, or convolutional neural network over word embeddings. For Task 2, they generated a  summary by selecting the sentences from the RP that are most relevant to the CPs using various features. They used CNN to learn the relation between a sentence and a  scoring value indicating its relevance.

\textbf{System 12:} The NLP-NITMZ team \cite{system_12} participated in all of the tasks (Tasks 1A, 1B and 2). For task  1A and 1B they extracted each citing papers (CP) text span that contains citations to the  reference paper (RP). They used cosine  similarity and Jaccard Similarity to measure the sentence similarity between CPs and RP, and picked the reference spans most similar to the citing sentence (Task 1A). For Task 1B, they applied rule based methods to extract the facets.  For Task 2, they built a summary generation system using the OpenNMT tool.

\textbf{System 20:} Team Magma \cite{system_20} treated Task 1A as a binary classification problem and explored several classifiers with different feature sets. They found that Logistic regression with  content-based features derived on topic and word similarities, in the ACL reference corpus, performed the best.

\section{Evaluation}

An automatic evaluation script was used to measure system performance for
\textbf{Task~1A}, in terms of the sentence ID overlaps between the sentences
identified in system output, versus the gold standard created by human annotators.
The raw number of overlapping sentences were used to calculate the precision,
recall and $F_1$ score for each system.

We followed the approach in most SemEval tasks in reporting the overall system performance as its micro-averaged performance over all topics in the blind test set. Additionally, we calculated lexical overlaps in terms of the ROUGE-2 and
ROUGE-SU4 scores~\cite{lin2004rouge} between the system
output and the human annotated gold standard reference spans. It should be noted that this year, the average performance on every task was obtained by calculating the average performance on each of three independent sets of annotations for Task 1, and the performance on the human summary was also an average of performances on three human summaries.

ROUGE scoring was used for Tasks~1a and Task~2.
Recall-Oriented Understudy for Gisting Evaluation
(ROUGE) is a set of metrics used to automatically evaluate
summarization systems~\cite{lin2004rouge} by measuring the overlap
between computer-generated summaries and multiple human written
reference summaries. ROUGE--2 measures the bigram overlap between the candidate
computer-generated summary and the reference summaries. More
generally, ROUGE--N measures the $n$-gram overlap. ROUGE-SU uses skip-bigram plus unigram overlaps. Similar to CL-SciSumm 2017, CL-SciSumm 2018 also uses ROUGE-2 and ROUGE-SU4 for its evaluation.\\

\textbf{Task~1B} was evaluated as a proportion of the correctly classified
discourse facets by the system, contingent on the expected response of
Task~1A. As it is a multi-label classification, this task was also
scored based on the precision, recall and $F_1$ scores.

\textbf{Task~2} was optional, and also evaluated using the ROUGE--2 and
ROUGE--SU4 scores between the system output and three types of gold
standard summaries of the research paper: the reference paper's
abstract, a community summary, and a human summary.

The evaluation scripts have been provided at the CL-SciSumm Github
repository\footnote{\url{github.com/WING-NUS/scisumm-corpus}} where
the participants may run their own evaluation and report the results.

\section{Results}
\label{s:sysruns}

This section compares the participating systems in terms
of their performance. Three of the ten systems that did
Task~1 also did the bonus Task~2. The results are provided in Table~\ref{tab:task1}
and Table ~\ref{tab:task2}. The detailed implementation of
the individual runs are described in the system papers included
in this proceedings volume.
\begin{table}[!ht]
\scalebox{0.9}{\begin{tabular}{|l|c|c|c|}
\hline
\textbf{System} & \shortstack{\textbf{Task 1A: Sentence}\\\textbf{Overlap ($F_1$)}} & \shortstack{\textbf{Task 1A:}\\\textbf{ROUGE $F_1$}} & \textbf{Task 1B}\\
\hline
system 6 Run 4           & 0.145                         & 0.131              & 0.262       \\
system 6 Run 2           & 0.139                         & 0.119              & 0.256       \\
system 6 Run 1           & 0.138                         & 0.117              & 0.257       \\
system 6 Run 3           & 0.128                         & 0.085              & 0.281       \\
system 2 Run 13          & 0.122                         & 0.049              & 0.261       \\
system 2 Run 12          & 0.122                         & 0.049              & 0.381       \\
system 2 Run 11          & 0.122                         & 0.049              & 0.313       \\
system 11 Voting3        & 0.117                         & 0.084              & 0.108       \\
system 2 Run 4           & 0.114                         & 0.053              & 0.341       \\
system 2 Run 6           & 0.114                         & 0.053              & 0.363       \\
system 2 Run 5           & 0.114                         & 0.053              & 0.277       \\
system 2 Run 7           & 0.114                         & 0.053              & 0.226       \\
system 9 KLBM25oR        & 0.114                         & 0.139              & 0.169       \\
system 2 Run 9           & 0.113                         & 0.052              & 0.356       \\
system 2 Run 8           & 0.113                         & 0.052              & 0.294       \\
system 2 Run 10          & 0.113                         & 0.052              & 0.256       \\
system 9 KLBM25noAuthoR  & 0.112                         & 0.089              & 0.236       \\
system 9 KLw2vWinIDFoRnE & 0.105                         & 0.067              & 0.288       \\
system 11 MJ1            & 0.099                         & 0.114              & 0.070       \\
system 12                & 0.094                         & 0.122              & 0.118       \\
system 7 Run 17          & 0.092                         & 0.053              & 0.302       \\
system 7 Run 19          & 0.091                         & 0.048              & 0.307       \\
system 7 Run 13          & 0.091                         & 0.069              & 0.245       \\
system 2 Run 1           & 0.090                         & 0.043              & 0.263       \\
system 2 Run 2           & 0.090                         & 0.043              & 0.299       \\
system 2 Run 3           & 0.090                         & 0.043              & 0.223       \\
system 11 BN1            & 0.089                         & 0.110              & 0.064       \\
system 7 Run 14          & 0.089                         & 0.075              & 0.225       \\
system 7 Run 20          & 0.087                         & 0.049              & 0.265       \\
system 7 Run 6           & 0.085                         & 0.070              & 0.200       \\
system 11 0.0001CNN4     & 0.083                         & 0.041              & 0.150       \\
system 7 Run 18          & 0.082                         & 0.047              & 0.266       \\
system 9 KLw2vIDFoRnE    & 0.082                         & 0.039              & 0.261       \\
system 7 Run 7           & 0.079                         & 0.072              & 0.182       \\
system 7 Run 4           & 0.079                         & 0.061              & 0.195       \\
system 7 Run 12          & 0.079                         & 0.035              & 0.330       \\
system 7 Run 11          & 0.079                         & 0.035              & 0.330       \\
system 7 Run 10          & 0.077                         & 0.046              & 0.224       \\
system 7 Run 3           & 0.076                         & 0.065              & 0.186       \\
system 7 Run 5           & 0.075                         & 0.060              & 0.196       \\
system 9 KLw2vWinoRnE    & 0.075                         & 0.070              & 0.168       \\
system 7 Run 9           & 0.071                         & 0.072              & 0.135       \\
system 11 Voting2        & 0.070                         & 0.025              & 0.122       \\
system 7 Run 2           & 0.070                         & 0.057              & 0.178       \\
system 4                 & 0.069                         & 0.027              & 0.169       \\
system 7 Run 8           & 0.068                         & 0.046              & 0.185       \\
system 7 Run 1           & 0.066                         & 0.051              & 0.180       \\
system 7 Run 16          & 0.059                         & 0.032              & 0.173       \\
system 9 KLw2vSIFoR      & 0.059                         & 0.039              & 0.171       \\
system 7 Run 15          & 0.049                         & 0.022              & 0.192       \\
system 20                & 0.049                         & 0.062              & 0.071       \\
system 10 Run 3          & 0.044                         & 0.035              & 0.132       \\
system 10 Run 2          & 0.042                         & 0.022              & 0.179       \\
system 10 Run 1          & 0.040                         & 0.021              & 0.175       \\
system 11 0.1CNN4        & 0.025                         & 0.023              & 0.083       \\
system 8 Run 3           & 0.023                         & 0.028              & 0.066       \\
system 8 Run 2           & 0.012                         & 0.016              & 0.050       \\
system 8 Run 1           & 0.011                         & 0.010              & 0.059       \\
system 8 Run 4           & 0.005                         & 0.010              & 0.022      \\
\hline
\end{tabular}}
\caption{Systems' performance in Task 1A and 1B, ordered by their
$F_1$-scores for sentence overlap on Task 1A.}
\label{tab:task1}
\end{table}

\begin{table}[!ht]
\renewcommand{\arraystretch}{1.15}
\centering
\small
\scalebox{0.85}{\begin{tabular}{|l | r | r | r | r | r | r |}
\hline
\multirow{2}{*}{\textbf{System}} &
\multicolumn{2}{c}{} \textbf{Vs. Abstract} & \multicolumn{2}{|c}{} \textbf{Vs. Community } &
\multicolumn{2}{|c|}{\textbf{Vs. Human} }
\\
\cline{2-7}
 & \textbf{R--2}& \textbf{RSU--4}& \textbf{R--2}& \textbf{RSU--4}& \textbf{R--2}& \textbf{RSU--4}\\
\hline
system 11 upf\_submission\_gar\_abstract     & 0.329        & 0.172          & 0.149         & 0.090           & 0.241     & 0.171       \\
system 11 upf\_submission\_sgar\_abstract    & 0.316        & 0.167          & 0.169         & 0.101           & 0.245     & 0.169       \\
system 11 upf\_submission\_rouge\_abstract   & 0.311        & 0.156          & 0.153         & 0.093           & 0.252     & 0.170       \\
system 11 upf\_submission\_acl\_abstract     & 0.245        & 0.145          & 0.130         & 0.083           & 0.173     & 0.142       \\
system 11 upf\_submission\_google\_abstract  & 0.230        & 0.128          & 0.129         & 0.077           & 0.172     & 0.125       \\
system 7 Run 4                               & 0.217        & 0.142          & 0.114         & 0.042           & 0.158     & 0.115       \\
system 2 Run 5                               & 0.215        & 0.115          & 0.138         & 0.074           & 0.220     & 0.151       \\
system 2 Run 11                              & 0.215        & 0.123          & 0.140         & 0.076           & 0.197     & 0.146       \\
system 7 Run 14                              & 0.210        & 0.120          & 0.087         & 0.027           & 0.135     & 0.086       \\
system 2 Run 13                              & 0.207        & 0.117          & 0.142         & 0.076           & 0.198     & 0.146       \\
system 2 Run 6                               & 0.207        & 0.118          & 0.135         & 0.073           & 0.204     & 0.144       \\
system 11 upf\_submission\_rouge\_human      & 0.204        & 0.108          & 0.147         & 0.084           & 0.197     & 0.146       \\
system 2 Run 12                              & 0.201        & 0.116          & 0.140         & 0.075           & 0.199     & 0.147       \\
system 2 Run 3                               & 0.199        & 0.113          & 0.133         & 0.071           & 0.220     & 0.152       \\
system 7 Run 3                               & 0.196        & 0.120          & 0.089         & 0.035           & 0.184     & 0.128       \\
system 2 Run 4                               & 0.195        & 0.113          & 0.146         & 0.074           & 0.220     & 0.155       \\
system 2 Run 2                               & 0.194        & 0.111          & 0.131         & 0.072           & 0.205     & 0.143       \\
system 11 upf\_submission\_gar\_human        & 0.193        & 0.095          & 0.123         & 0.076           & 0.208     & 0.144       \\
system 2 Run 1                               & 0.193        & 0.112          & 0.137         & 0.073           & 0.215     & 0.153       \\
system 2 Run 7                               & 0.193        & 0.113          & 0.135         & 0.076           & 0.207     & 0.146       \\
system 7 Run 1                               & 0.189        & 0.104          & 0.091         & 0.034           & 0.147     & 0.090       \\
system 11 upf\_submission\_sgar\_human       & 0.187        & 0.095          & 0.124         & 0.075           & 0.191     & 0.135       \\
system 12                                    & 0.185        & 0.110          & 0.213         & 0.107           & 0.217     & 0.153       \\
system 7 Run 16                              & 0.183        & 0.106          & 0.080         & 0.032           & 0.140     & 0.095       \\
system 11 upf\_submission\_rouge\_community  & 0.181        & 0.099          & 0.148         & 0.092           & 0.187     & 0.133       \\
system 7 Run 12                              & 0.179        & 0.108          & 0.088         & 0.025           & 0.116     & 0.073       \\
system 7 Run 11                              & 0.179        & 0.108          & 0.088         & 0.025           & 0.116     & 0.073       \\
system 7 Run 17                              & 0.179        & 0.109          & 0.095         & 0.032           & 0.135     & 0.083       \\
system 7 Run 13                              & 0.176        & 0.106          & 0.090         & 0.035           & 0.139     & 0.097       \\
system 7 Run 15                              & 0.173        & 0.105          & 0.093         & 0.035           & 0.145     & 0.097       \\
system 7 Run 10                              & 0.172        & 0.102          & 0.067         & 0.025           & 0.114     & 0.075       \\
system 7 Run 6                               & 0.171        & 0.099          & 0.083         & 0.030           & 0.134     & 0.094       \\
system 7 Run 19                              & 0.170        & 0.095          & 0.082         & 0.028           & 0.131     & 0.078       \\
system 11 upf\_submission\_gar\_community    & 0.168        & 0.094          & 0.154         & 0.093           & 0.156     & 0.123       \\
system 7 Run 5                               & 0.166        & 0.099          & 0.104         & 0.039           & 0.137     & 0.106       \\
system 7 Run 7                               & 0.163        & 0.112          & 0.102         & 0.037           & 0.136     & 0.102       \\
system 11 upf\_submission\_summa\_abstract   & 0.162        & 0.081          & 0.154         & 0.090           & 0.189     & 0.138       \\
system 2 Run 10                              & 0.158        & 0.081          & 0.104         & 0.058           & 0.113     & 0.084       \\
system 7 Run 18                              & 0.156        & 0.089          & 0.085         & 0.029           & 0.135     & 0.091       \\
system 11 upf\_submission\_sgar\_community   & 0.155        & 0.084          & 0.159         & 0.096           & 0.148     & 0.116       \\
system 7 Run 2                               & 0.153        & 0.094          & 0.087         & 0.027           & 0.131     & 0.083       \\
system 7 Run 20                              & 0.152        & 0.090          & 0.085         & 0.032           & 0.123     & 0.086       \\
system 11 upf\_submission\_acl\_human        & 0.141        & 0.070          & 0.106         & 0.066           & 0.148     & 0.109       \\
system 11 upf\_submission\_google\_human     & 0.137        & 0.067          & 0.101         & 0.065           & 0.134     & 0.100       \\
system 11 upf\_submission\_acl\_community    & 0.134        & 0.079          & 0.129         & 0.083           & 0.119     & 0.105       \\
system 7 Run 8                               & 0.129        & 0.071          & 0.085         & 0.025           & 0.111     & 0.069       \\
system 2 Run 9                               & 0.119        & 0.077          & 0.108         & 0.055           & 0.086     & 0.085       \\
system 11 upf\_submission\_summa\_human      & 0.118        & 0.063          & 0.117         & 0.072           & 0.143     & 0.105       \\
system 11 upf\_submission\_google\_community & 0.110        & 0.071          & 0.116         & 0.072           & 0.124     & 0.109       \\
system 2 Run 8                               & 0.109        & 0.073          & 0.122         & 0.058           & 0.071     & 0.070       \\
system 7 Run 9                               & 0.101        & 0.071          & 0.088         & 0.031           & 0.131     & 0.082       \\
system 11 upf\_submission\_summa\_community  & 0.096        & 0.053          & 0.134         & 0.078           & 0.138     & 0.096     \\
\hline
\end{tabular}}
\caption{Systems' performance for Task 2 ordered by their
ROUGE--2(R--2) $F_1$-scores.}
\label{tab:task2}
\end{table}

For Task~1A, on using sentence overlap (F1 score) as the metric, the best performance was by four runs from NUDT (system 6) \cite{system_6}. Their performance was closely followed by three runs from CIST (system 2) \cite{system_2}. The third best system was UPF-TALN (system 11) \cite{system_11}.
When ROUGE-based F1 is used as a metric, the best performance is by Klick Labs (system 9) \cite{system_9} 
followed by NUDT (system 6) \cite{system_6} 
and then NLP-NITMZ (system 12) \cite{system_12}. 

The best performance in Task~1B was by several runs submitted by CIST (system 2) \cite{system_2} followed by NJUST (system 7) \cite{system_7}. Klick Labs (system 9) \cite{system_9} was the second runner-up.

For Task~2, TALN-UPF (system 11) \cite{system_11} had the best performance against the abstract and human summaries, and the second-best performance against community summaries. NLP-NITMZ (system 12) \cite{system_12} had the best performance against the community summaries and were the second runners-up in the evaluation against human summaries. CIST (system 2) \cite{system_2} summaries had the second best performance against human summaries and finished as second runners-up against abstract and community summaries.



\section{Conclusion and Future Work}
\label{s:conc}

Ten teams participated in this year's shared task, on a corpus that was 33\% larger than the 2017 corpus. In follow-up work, we plan to release a detailed comparison of the annotations as well as a micro-level error analysis to identify possible gaps in document or annotation quality. We will also aim to expand the part of the corpus with multiple annotations, in the coming few months. Furthermore, we expect to release other resources complementary to the CL-scientific summarization task, such as semantic concepts from the ACL Anthology
Network~\cite{radev2009acl}.

We believe that the large improvements in Task~1A this year are a sign of forthcoming breakthroughs in information retrieval and summarization methods, and we hope that the community will not give up on the challenging task of generating scientific summaries for computational linguistics. Based on the experience of running this task for four years, we believe that lexical methods would work well with the structural and semantic characteristics that are unique to scientific
documents, and perhaps will be complemented with domain-specific word embeddings in a deep learning framework. The Shared Task has demonstrated potential as a transfer learning task~\cite{conroy15:_vector} and is also expected to allow the generalization of its methods to other areas of scientific summarization.

\renewcommand{\abstractname}{\ackname}
\begin{abstract}

We would like to thank Microsoft Research Asia, for their generous
funding. We would also like to thank Vasudeva Varma and colleagues at
IIIT-Hyderabad, India and University of Hyderabad for their efforts in
convening and organizing our annotation workshops. We are grateful to our hardworking and talented annotators: Akansha Gehlot, Ankita Patel, Fathima Vardha, Swastika Bhattacharya  and Sweta Kumari, without whom the Cl-SciSumm corpus -- and ultimately the Shared Task itself -- would not have been possible. We would also like to thank Rahul Jha and Dragomir Radev for sharing their software to prepare the XML versions of papers and constructing the test corpus for 2018. Finally, we acknowledge the continued advice of Hoa Dang, Lucy Vanderwende and Anita de Waard from the pilot stage of this task. We are grateful to Kevin B. Cohen and colleagues for their support, and for sharing their annotation schema, export scripts and the Knowtator package implementation on the Protege software -- all of which have been indispensable for this shared task.
\end{abstract}


%

\bibliographystyle{splncs03}
\bibliography{splncs}

\begin{thebibliography}{10}
\providecommand{\url}[1]{\texttt{#1}}
\providecommand{\urlprefix}{URL }

\bibitem{system_11}
Abura'ed, A., Bravo, A., Chiruzzo, L., Saggion1, H.: {LaSTUS/TALN+INCO @
  CL-SciSumm 2018 - Using Regression and Convolutions for Cross-document
  Semantic Linking and Summarization of Scholarly Literature}. In: Proceedings
  of the 3nd Joint Workshop on Bibliometric-enhanced Information Retrieval and
  Natural Language Processing for Digital Libraries {(BIRNDL2018)}. Ann Arbor,
  Michigan (July 2018)

\bibitem{system_8}
Agrawal, K., Mittal, A.: {IIIT-H @ CLScisumm-18}. In: Proceedings of the 3nd
  Joint Workshop on Bibliometric-enhanced Information Retrieval and Natural
  Language Processing for Digital Libraries {(BIRNDL2018)}. Ann Arbor, Michigan
  (July 2018)

\bibitem{system_20}
Alonso, H.M., Makki, R., Gu, J.: {CL-SciSumm Shared Task - Team Magma}. In:
  Proceedings of the 3nd Joint Workshop on Bibliometric-enhanced Information
  Retrieval and Natural Language Processing for Digital Libraries
  {(BIRNDL2018)}. Ann Arbor, Michigan (July 2018)

\bibitem{system_9}
Baruah, G., Kolla, M.: {Klick Labs at CL-SciSumm 2018}. In: Proceedings of the
  3nd Joint Workshop on Bibliometric-enhanced Information Retrieval and Natural
  Language Processing for Digital Libraries {(BIRNDL2018)}. Ann Arbor, Michigan
  (July 2018)

\bibitem{conroy15:_vector}
Conroy, J., Davis, S.: Vector space and language models for scientific document
  summarization. In: {NAACL-HLT}. pp. 186--191. Association of Computational
  Linguistics, Newark, NJ, USA (2015)

\bibitem{system_4}
Davoodi, E., Madan, K., Gu, J.: {CLSciSumm Shared Task: On the Contribution of
  Similarity measure and Natural Language Processing Features for Citing
  Problem}. In: Proceedings of the 3nd Joint Workshop on Bibliometric-enhanced
  Information Retrieval and Natural Language Processing for Digital Libraries
  {(BIRNDL2018)}. Ann Arbor, Michigan (July 2018)

\bibitem{system_12}
Debnath, D., Achom, A., Pakray, P.: {NLP-NITMZ @ CLScisumm-18}. In: Proceedings
  of the 3nd Joint Workshop on Bibliometric-enhanced Information Retrieval and
  Natural Language Processing for Digital Libraries {(BIRNDL2018)}. Ann Arbor,
  Michigan (July 2018)

\bibitem{jaidka2017overview}
Jaidka, K., Chandrasekaran, M.K., Jain, D., Kan, M.Y.: The cl-scisumm shared
  task 2018: Results and key insights. In: Proceedings of the 2nd Joint
  Workshop on Bibliometric-enhanced Information Retrieval and Natural Language
  Processing for Digital Libraries {(BIRNDL} 2017) co-located with the 40th
  International {ACM} {SIGIR} Conference on Research and Development in
  Information Retrieval {(SIGIR} 2017), Tokyo, Japan, August 11, 2017. (2017),
  \url{http://ceur-ws.org/Vol-1888/editorial.pdf}

\bibitem{jaidka2017insights}
Jaidka, K., Chandrasekaran, M.K., Rustagi, S., Kan, M.Y.: Insights from
  cl-scisumm 2016: the faceted scientific document summarization shared task.
  International Journal on Digital Libraries pp. 1--9 (2017)

\bibitem{jaidka2013deconstructing}
Jaidka, K., Khoo, C.S., Na, J.C.: Deconstructing human literature reviews--a
  framework for multi-document summarization. In: {Proc. of ENLG}. pp. 125--135
  (2013)

\bibitem{sparckjones2007automatic}
Jones, K.S.: {Automatic summarising: The state of the art}. Information
  Processing and Management  43(6),  1449--1481 (2007)

\bibitem{lin2004rouge}
Lin, C.Y.: Rouge: A package for automatic evaluation of summaries. {Text
  summarization branches out: Proceedings of the ACL-04 workshop}  8 (2004)

\bibitem{system_2}
Ma, S., Xu, J., Wang, J., Zhang, C.: {NJUST@CLSciSumm-17}. In: Proc.\ of the
  2nd Joint Workshop on Bibliometric-enhanced Information Retrieval and Natural
  Language Processing for Digital Libraries {(BIRNDL2017)}. Tokyo, Japan
  (August 2017)

\bibitem{system_7}
Ma, S., Zhang, H., Xu, J., Zhang, C.: {NJUST @ CLSciSumm-18}. In: Proceedings
  of the 3nd Joint Workshop on Bibliometric-enhanced Information Retrieval and
  Natural Language Processing for Digital Libraries {(BIRNDL2018)}. Ann Arbor,
  Michigan (July 2018)

\bibitem{MayrCJ17}
Mayr, P., Chandrasekaran, M.K., Jaidka, K.: Editorial for the 2nd joint
  workshop on bibliometric-enhanced information retrieval and natural language
  processing for digital libraries {(BIRNDL)} at {SIGIR} 2017. In: Proceedings
  of the 2nd Joint Workshop on Bibliometric-enhanced Information Retrieval and
  Natural Language Processing for Digital Libraries {(BIRNDL} 2017) co-located
  with the 40th International {ACM} {SIGIR} Conference on Research and
  Development in Information Retrieval {(SIGIR} 2017), Tokyo, Japan, August 11,
  2017. pp. 1--6 (2017), \url{http://ceur-ws.org/Vol-1888/editorial.pdf}

\bibitem{mayr16:_editorial}
Mayr, P., Frommholz, I., Cabanac, G., Wolfram, D.: {Editorial for the Joint
  Workshop on Bibliometric-enhanced Information Retrieval and Natural Language
  Processing for Digital Libraries (BIRNDL) at JCDL 2016}. In: Proc.\ of the
  Joint Workshop on Bibliometric-enhanced Information Retrieval and Natural
  Language Processing for Digital Libraries {(BIRNDL2016)}. pp. 1--5. Newark,
  NJ, USA (June 2016)

\bibitem{system_10}
Moraes, L.F.D., Das, A., Karimi, S., Verma, R.: {University of Houston @
  CL-SciSumm 2018}. In: Proceedings of the 3nd Joint Workshop on
  Bibliometric-enhanced Information Retrieval and Natural Language Processing
  for Digital Libraries {(BIRNDL2018)}. Ann Arbor, Michigan (July 2018)

\bibitem{nakov2004citances}
Nakov, P.I., Schwartz, A.S., Hearst, M.: {Citances: Citation sentences for
  semantic analysis of bioscience text}. In: {Proceedings of the SIGIR'04
  workshop on Search and Discovery in Bioinformatics}. pp. 81--88 (2004)

\bibitem{qazvinian2008scientific}
Qazvinian, V., Radev, D.: Scientific paper summarization using citation summary
  networks. In: {Proceedings of the 22nd International Conference on
  Computational Linguistics-Volume 1}. pp. 689--696. ACL (2008)

\bibitem{radev2009acl}
Radev, D.R., Muthukrishnan, P., Qazvinian, V.: The acl anthology network
  corpus. In: Proceedings of the 2009 Workshop on Text and Citation Analysis
  for Scholarly Digital Libraries. pp. 54--61. Association for Computational
  Linguistics (2009)

\bibitem{system_6}
Wang, P., Li, S., Wang, T., Zhou, H., Tang, J.: {NUDT @ CLSciSumm-18}. In:
  Proceedings of the 3nd Joint Workshop on Bibliometric-enhanced Information
  Retrieval and Natural Language Processing for Digital Libraries
  {(BIRNDL2018)}. Ann Arbor, Michigan (July 2018)

\end{thebibliography}
\end{document}